\documentclass[acmsmall,screen]{acmart}
\usepackage{multirow}
\AtBeginDocument{%
  }

\acmJournal{CSUR}
\acmVolume{58}
\acmNumber{8}
\acmArticle{215}
\acmMonth{02}
\acmDOI{10.1145/3789495}
\acmBadgeR[https://hulianyu.xyz/Interpretable-Clustering-Repository/]{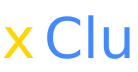}
\microtypesetup{expansion=false}   

\makeatletter
\AtBeginDocument{\let\MakeTextUppercase\@firstofone}
\def\@authorfont{\large\sffamily}
\AtBeginDocument{\def\@permissionCodeTwo{0360-0300}}
\AtBeginDocument{%
	\def\@ICRfoot{\footnotesize \@journalNameShort, Vol.~\@acmVolume, No.~\@acmNumber,
		Article~\@acmArticle, \@acmYear}%
	\fancypagestyle{standardpagestyle}{%
		\fancyhf{}%
		\renewcommand{\headrulewidth}{\z@}\renewcommand{\footrulewidth}{\z@}%
		\def\@acmArticlePage{\ifx\@acmArticle\empty\if@ACM@printfolios\thepage\fi
			\else\@acmArticle\if@ACM@printfolios:\thepage\fi\fi}%
		\fancyhead[LE]{\ACM@linecountL\@headfootfont\@acmArticlePage}%
		\fancyhead[RO]{\@headfootfont\@acmArticlePage}%
		\fancyhead[RE]{\@headfootfont\@shortauthors}%
		\fancyhead[LO]{\ACM@linecountL\@headfootfont\shorttitle}%
		\fancyfoot[RO,LE]{\@ICRfoot}%
	}%
	\fancypagestyle{firstpagestyle}{%
		\fancyhf{}%
		\renewcommand{\headrulewidth}{\z@}\renewcommand{\footrulewidth}{\z@}%
		\fancyfoot[RO,LE]{\@ICRfoot}%
		\fancyhead[LE]{\ACM@linecountL\@acmBadgeL}\fancyhead[LO]{\ACM@linecountL\@acmBadgeL}%
		\fancyhead[RO]{\@acmBadgeR}\fancyhead[RE]{\@acmBadgeR}%
	}%
	\pagestyle{standardpagestyle}%
}
\makeatother
\begin{document}

\title{Interpretable Clustering: A Survey}
\titlenote{This is the authors' version of the work, made available for personal and non-commercial use. The definitive Version of Record was published in \emph{ACM Computing Surveys}, Vol.~58, No.~8, Article~215 (2026), \url{https://doi.org/10.1145/3789495}.}

\author{Lianyu Hu}
\email{hly4ml@gmail.com}
\orcid{0000-0001-7470-9395}
\affiliation{%
	\institution{College of Information Science and Engineering, Henan University of Technology}
	\city{Zhengzhou}
	\state{Henan}
	\country{China}
}

\author{Mudi Jiang}
\affiliation{%
	\institution{School of Software, Dalian University of Technology}
	\city{Dalian}
	\state{Liaoning}
	\country{China}
}

\author{Junjie Dong}
\affiliation{%
	\institution{Xinchang Power Supply Company, State Grid Corporation of China}
	\city{Shaoxing}
	\state{Zhejiang}
	\country{China}
}

\author{Xinying Liu}
\affiliation{%
	\institution{School of Software, Dalian University of Technology}
	\city{Dalian}
	\state{Liaoning}
	\country{China}
}

\author{Zengyou He}
\email{zyhe@dlut.edu.cn}
\affiliation{%
	\institution{School of Software, Dalian University of Technology}
	\city{Dalian}
	\state{Liaoning}
	\country{China}
}

\renewcommand{\shortauthors}{L. Hu et al.}

\begin{abstract}
In recent years, much of the research on clustering algorithms has primarily focused on enhancing their accuracy and efficiency, frequently at the expense of interpretability. However, as these methods are increasingly being applied in high-stakes domains such as healthcare, finance, and autonomous systems, the need of transparent and interpretable clustering outcomes has become a critical concern. This is not only necessary for gaining user trust but also for satisfying the growing ethical and regulatory demands in these fields. Ensuring that decisions derived from clustering algorithms can be clearly understood and justified is now a fundamental requirement. To address this need, this paper provides a comprehensive and structured review of the current state of explainable clustering algorithms, identifying key criteria to distinguish between various methods. These insights can effectively assist researchers in making informed decisions about the most suitable explainable clustering methods for specific application contexts, while also promoting the development and adoption of clustering algorithms that are both efficient and transparent. For convenient access and reference, an open repository organizes representative and emerging interpretable clustering methods under the taxonomy proposed in this survey, available at \url{https://hulianyu.xyz/Interpretable-Clustering-Repository/}
\end{abstract}




\makeatletter\@printpermissionfalse\makeatother
\maketitle
\begingroup
\renewcommand{\thefootnote}{}
\footnotetext{Lianyu Hu and Mudi Jiang both contributed equally to this research. Zengyou He is the corresponding author.}
\endgroup

\section{Introduction}
Cluster analysis \cite{Jain2010,saxena2017review} is a crucial task in the field of data mining, which aims to partition data into distinct groups based on the intrinsic characteristics and patterns within the data. This process helps in uncovering meaningful structures and relationships among data points, facilitating various applications and further analysis.

For decades, numerous algorithms have been proposed to solve clustering problems across different applications, achieving high accuracy. However, in most cases, clustering models exist as black boxes, leading to common questions such as: How are the clustering results formed? Can people understand the logic behind the formation of the clustering results? Is the model trustworthy? The clustering model's ability to explain such issues is tentatively defined as model's clustering interpretability or explainability \cite{bertsimas2018interpretable}. Explainability is commonly defined as the extent to which the internal mechanics of a machine learning system can be clarified in human terms, i.e., explanations should be faithful to the model while providing information that is relevant in the current context~\cite{Molnar2025}. Given that most researchers in data mining and machine learning use interpretability  and explainability interchangeably~\cite{Atzmueller2024}, this paper will use the term interpretability throughout this paper.

To date, interpretability still lacks a precise or mathematical definition~\cite{Murdoch2019,Allen2023,Krishna2024}. Different sources provide slightly varying definitions -- for instance, it is defined as ``the ability to explain or to present in understandable terms to a human'' in \cite{doshi2017towards}, ``the degree to which a human can understand the cause of a decision" in \cite{miller2019explanation}, and ``make the behavior and predictions of machine learning systems understandable to humans" in \cite{Molnar2025}. Collectively, these definitions can all capture the essence of interpretability. 

However, the interpretability of a model may vary depending on the user's actual needs and can manifest in different dimensions. In studies of specific diseases, physicians are often more concerned with identifying patient characteristics that indicate a higher likelihood of having the disease and whether these characteristics can assist in early diagnosis. In contrast, data scientists focus on designing interpretable models that provide compelling explanations for patients and effectively elucidate the reasons behind each patient's assignment to a particular disease type, thereby aiding in understanding the impact of various characteristics on the outcomes. Therefore, although various interpretable methods can provide different degrees of interpretability across multiple  dimensions, it remains necessary to provide a systematic summary and distinction of these methods.

As far as we know, there have been several reviews that summarize methods related to interpretability. However, most existing reviews are either too general~\cite{Ullah2024,Ali2023,Bodria2023,Guidotti2018} or focus on specific domains~\cite{Xu2025,Glanois2024,Sabbatini2026,li2023survey}, and do not focus on the clustering domain. Moreover, the only survey closely related to this topic~\cite{yang2021survey} was published relatively early and thus does not include the latest research or emerging ideas in this field. To fill this gap, we have comprehensively collected existing interpretable clustering methods and proposed a set of criteria to classify them, ensuring that all methods related to interpretable clustering can be categorized under one of these criteria. Furthermore, we divide the clustering process into three stages and classify all interpretable clustering methods according to their interpretability at different stages, providing the overall framework for this review: (1) the feature selection stage (pre-clustering), (2) the model building stage (in-clustering), and (3) the model explanation stage (post-clustering). We believe this review will provide readers with a new understanding of interpretable clustering and lay a foundation for future research in this area.

The rest of this paper is organized as follows. Section \ref{sec:need} discusses the need of interpretable clustering. Section \ref{sec:taxonomy} provides a taxonomy of interpretable clustering methods. Sections \ref{sec:pre} to \ref{sec:post} review interpretable pre-clustering, in-clustering, and post-clustering methods, respectively, based on different stages of interpretability in the clustering process. Finally, Section \ref{conclusion} concludes the paper and discusses future directions.

\section{The need for interpretable clustering}
\label{sec:need}
As artificial intelligence and machine learning algorithms become more advanced and excel in various tasks, they are increasingly being applied across multiple domains. However, their use remains limited in risk-sensitive areas such as healthcare, justice, manufacturing, defense, and finance. The application of AI systems and the underlying machine learning algorithms in these fields involves three key human roles~\cite{Gunning2019}: developers, end users within the relevant domain, and regulators at the societal level. For any of these roles, it is crucial for humans to understand and trust how the algorithm arrives at its results. For instance, developers need to understand how the algorithm produces meaningful outcomes and recognize its limitations, enabling them to correct errors or conduct further assessments. End users need to evaluate whether the algorithm's results incorporate domain-specific knowledge and are well-founded. Regulators need to consider the implications of the algorithm's outcomes, such as fairness, potential discrimination, and where the risks and responsibilities lie. This necessitates transparency and trustworthiness~\cite{Wang2023} throughout the entire algorithmic process.

In response to these challenges, research in interpretable machine learning has gained momentum~\cite{Molnar2025}. Much of the downstream analysis is typically built at the cluster level, where clustering methods are designed to generate patterns as the initial understanding of the data. At this stage, the need for interpretability of clustering, along with the transparency of algorithmic mechanisms, becomes increasingly pronounced.

\subsection{What is interpretable clustering?}
Conventional clustering algorithms typically focus on delivering clustering results, treating accuracy and efficiency as top priorities, especially in complex, high-dimensional data. The models they employ are largely ``black boxes'', particularly in the case of advanced clustering methods that often utilize representation learning techniques and deep learning. These methods consider all dimensions and feature values of the data, actively involving them in the generation of clustering results. However, the reasoning behind ``why'' and ``how'' these results are generated remains opaque to the algorithm designers, making it even more difficult for end users to comprehend. In contrast, interpretable clustering methods explicitly aim to explain the clustering results, enabling humans to understand why the algorithmic process produces meaningful clustering outcomes. 

For instance, in a loan evaluation scenario, a conventional clustering algorithm may integrate all applicant attributes such as age, gender, income, and credit score into its objective function, treating them as equally important in forming clusters. While this may produce well-separated applicant groups, it remains unclear which attributes actually drive these distinctions. In contrast, an interpretable clustering method can reveal the decisive factors such as low credit scores or high debt-to-income ratios that define a cluster, allowing the reasoning behind the grouping to be clearly understood and validated.

Interpretable clustering, in general, aims to make such reasoning explicit by incorporating mechanisms that expose how cluster assignments are determined. Any approach that provides interpretability during the clustering process or makes the resulting outcomes easier to understand can be categorized under this domain. A hallmark of these methods is the integration of interpretable models~\cite{Rudin2019} at various stages of the clustering pipeline. These interpretable components accompany the final clustering results, making them understandable, trustworthy~\cite{Han2025}, and usable by humans. Such components may include, but are not limited to, the use of specific feature values (e.g., age, income) combined with explicit model syntax (e.g., rules, trees) to identify key factors that contribute to the clustering outcomes, allowing end users to comprehend the results and validate the conclusions derived from these interpretable elements.

\subsection{What is a good interpretable clustering method?}
An interpretable clustering method provides clear evidence to explain how clustering results are derived, offering end users the opportunity to understand both the behavior of the algorithm and the logic behind the clustering outcomes. However, whether end users ultimately choose to trust this evidence may depend on application-driven needs or expert knowledge. As machine learning researchers and data scientists, we are primarily equipped to assess what constitutes a good interpretable clustering method from a data-driven perspective. 

A good interpretable clustering method can therefore be characterized by the following key properties, reflecting its ability to provide explanations in two complementary aspects:
\begin{itemize}
	\item \textbf{Simplicity and parsimony.} The form of interpretable evidence should be as concise as possible, minimizing the number of feature values or conditions used to define each cluster. A simpler explanatory form not only reduces the cognitive load on end users but also allows them to follow the reasoning process more easily. In other words, the fewer features required to describe how a cluster is formed, the more transparent and accessible the explanation becomes.
	\item \textbf{Uniqueness and exclusivity.} Each cluster should convey distinct and non-overlapping information, reflecting its own functional role within the data. Ideally, the same interpretable evidence should correspond to only one specific cluster, ensuring that the explanation remains both credible and unambiguous. This exclusivity enhances trust, as end users can be confident that the evidence is tightly linked to a single, well-defined cluster, rather than ambiguously shared among multiple clusters. Overlapping explanations across clusters would blur these distinctions and risk confusion or misinterpretation.
\end{itemize}

These principles can further guide how the quality of interpretability is evaluated in practice. To determine the goodness of an interpretable clustering method, or even to quantify it, one must consider the specific interpretable model being used. For example, when utilizing decision tree models, it is clear that the evidence used to define each cluster is highly distinctive through the tree's splits, thereby satisfying the basic requirement of uniqueness. Additionally, one can measure how easily end users understand the results by examining the structural parameters of the tree~\cite{Piltaver2016}, such as the number of leaf nodes (i.e., the number of clusters) and the average depth of the tree. The process from data to clusters is represented by paths from the root to the leaf nodes, with each branching node recording the decision (splitting feature value) that leads to a cluster. Using fewer feature values results in more concise interpretable evidence, making it easier for end users to understand and trust the clustering results.

\begin{figure*}[t]
	\centering
	\includegraphics[width=\linewidth]{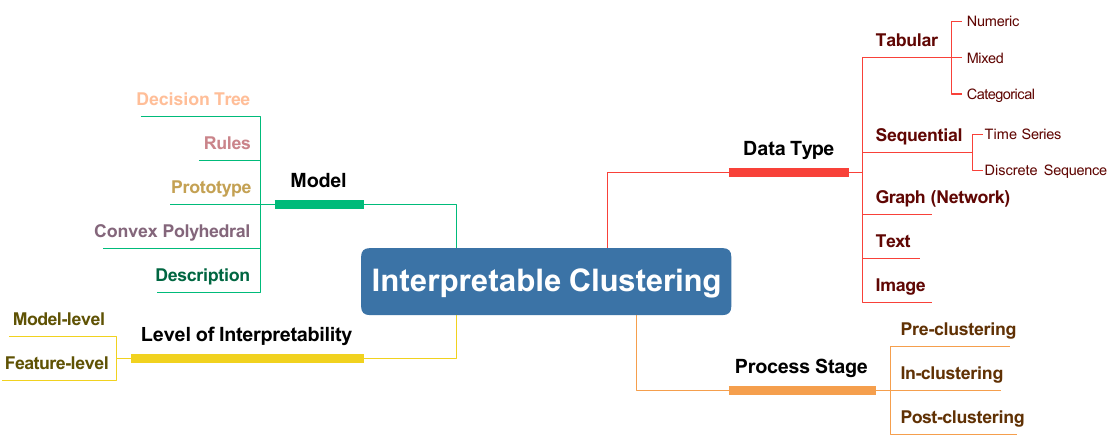}
	\caption{Interpretable clustering taxonomy categorized by distinct criteria, most existing methods align with a single category per criterion.}  
	\label{taxonomy}   
\end{figure*}

\section{A taxonomy of interpretable clustering methods}
\label{sec:taxonomy}

In this section, after collecting and systematically reviewing existing interpretable clustering methods, we establish a unified taxonomy that organizes them along four distinct criteria, as illustrated in Fig.~\ref{taxonomy}. This taxonomy provides a structured framework for classifying interpretable clustering methods according to four complementary dimensions: \textbf{(1) Process stage}, \textbf{(2) Interpretable model}, \textbf{(3) Interpretability level}, and \textbf{(4) Data modality}. Together, these dimensions comprehensively characterize how existing approaches achieve interpretability and serve as a foundation for identifying appropriate methods under different analytical contexts and requirements.

\textbf{(1) Process stage.} Based on widely recognized clustering processes, interpretable clustering methods can be categorized into three types: \textit{pre-clustering}, \textit{in-clustering}, and \textit{post-clustering}. \textit{Pre-clustering} methods are typically executed before the clustering process and focus on selecting or extracting interpretable features, aiming to provide human-understandable representations for subsequent clustering. \textit{In-clustering} methods integrate interpretability directly into the model construction process, generating accurate partitions whose formation mechanisms are inherently transparent, without the need for additional post-hoc explanations. \textit{Post-clustering} methods, in contrast, interpret the results of existing black-box clustering models by constructing surrogate interpretable models that reveal the reasoning behind previously opaque clustering outcomes.

\begin{figure*}[t]
	\centering
	\includegraphics[width=\linewidth]{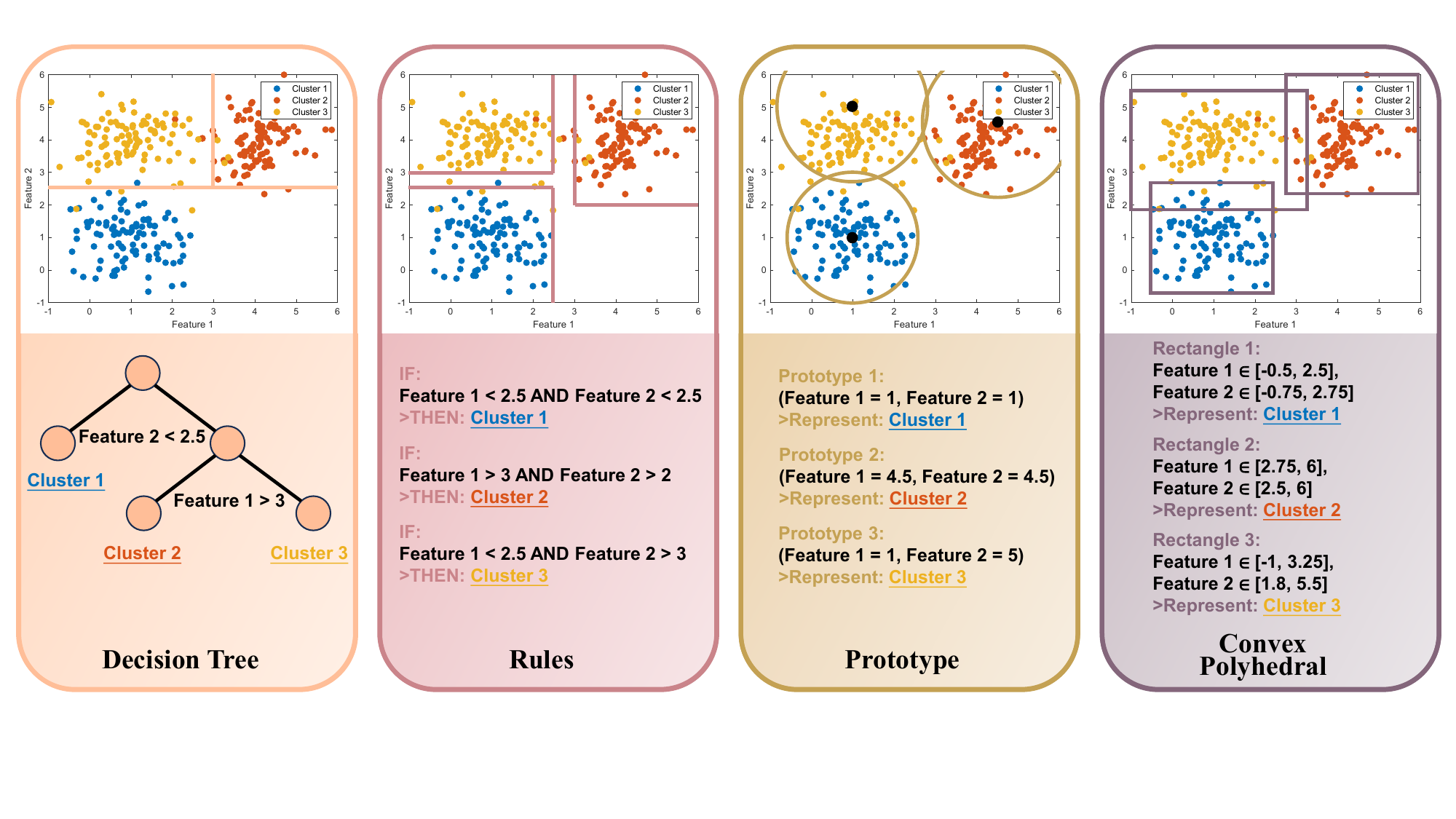}
	\caption{Illustration of four interpretable clustering models applied to the same two-dimensional dataset with three Gaussian clusters. The upper panels display how each model partitions the feature space, while the bottom panels show the feature values used for interpretability.}  
	\label{models}   
\end{figure*}

\textbf{(2) Interpretable model.} A large proportion of in-clustering and post-clustering approaches can be distinguished by the type of interpretable model they employ. As illustrated in Fig.~\ref{models}, four representative model families are commonly used to construct explainable cluster structures, each offering a distinct mechanism for generating human-understandable evidence of how clusters are formed.

\begin{itemize}
	\item \textbf{\textit{Decision tree.}} The decision tree model is one of the most widely recognized interpretable models in machine learning and has long been used for classification and regression tasks.  
	Its interpretability stems from the recursive, hierarchical splitting of data based on feature values to generate intermediate results, where the final outputs can be traced through the sequence of splits.  
	Instances are allocated to different leaf nodes (clusters) according to specific splitting criteria, following a transparent path from the root node, which represents the entire dataset, down through the branching nodes.  
	This hierarchical structure allows end users to understand both \textit{why} and \textit{how} each sample is assigned to a cluster, making the reasoning process explicit and verifiable.
	
	\item \textbf{\textit{Rules.}} In contrast to decision tree-based models, where users must trace a hierarchical path from the root to a leaf node, rule-based methods provide a more direct way to understand how clusters are defined. Interpretability arises from generating a set of candidate rules based on feature values, typically expressed as logical combinations (conjunctions or disjunctions) of conditions at the same level, such as ``age $>$ 40 and income $<$ 30k.'' These flat, non-hierarchical structures describe cluster-defining conditions in a way that is both concise and intuitive for end users, avoiding the progressive complexity that can arise in deep trees.
	
	\item \textbf{\textit{Prototype.}} The concept of a prototype, also referred to as an exemplar, can be understood analogously to the centroid in the $k$-means algorithm. Each prototype serves as a representative of its corresponding cluster, and samples that are sufficiently close to a given prototype are considered members of that cluster. Interpretability in this case arises from identifying which representative instance or exemplar contributes most to defining the cluster. Unlike hierarchical or rule-based models, prototype-based approaches emphasize \textit{representativeness} rather than logical structure, and may allow a certain degree of overlap among the clusters being explained.
	
	\item \textbf{\textit{Convex polyhedral.}} A convex polyhedral model generalizes the geometric intuition of convex polygons to higher dimensions, where each cluster is represented as a convex region enclosed by bounding planes formed by the intersection of a finite number of half-spaces. Interpretability arises from the explicit geometric constraints that delineate each region, allowing cluster membership to be visually and spatially verified.  Unlike prototype-based models with boundaries implicitly defined by similarity measures, convex polyhedral models provide explicit and well-defined geometric boundaries.
	
	\item \textbf{\textit{Description-based.}} Beyond the four canonical forms, data-type-dependent interpretable models exist that do not fit neatly into a single category. A representative example is the description-based approach, where clusters are defined by concise, human-readable predicates or summaries derived directly from data semantics. Unlike decision tree or rule-based models with fixed syntactic forms, or geometric models such as prototype- and polyhedral-based methods that rely on spatial continuity, description-based models are particularly common in relational or graph-structured settings. In community analysis, for instance, explanations often rely on node attributes, structural connectivity, or semantic relations, providing non-vector, structurally grounded summaries that capture each community's distinctive role and clearly distinguish it from other communities.
	
\end{itemize}

\textbf{\textit{Illustrative example.}} Consider a two-dimensional dataset with two features $(F_1, F_2)$ and three natural clusters, as illustrated in Fig.~\ref{models}. This illustrative example is used to show how different interpretable models provide transparent reasoning about the formation of clusters.

\begin{itemize}

\item \textbf{Decision tree.} The decision tree partitions the feature space through a sequence of hierarchical splits. For example, the model may first divide the data based on whether $F_2 < 2.5$, forming Cluster~1 for samples that satisfy this condition (left branch). For the remaining samples with $F_2 \geq 2.5$, a second split on $F_1 > 3$ further separates them into Cluster~2 and Cluster~3 ($F_1 \leq 3$). Each cluster can thus be traced through an explicit decision path from root to leaf. For instance, the path ``$F_2 \geq 2.5$ (at the root node) $\rightarrow$ $F_1 > 3$ (at the right branch)'' corresponds to Cluster~2, clearly illustrating the sequential reasoning behind the assignment.

\item \textbf{Rule.} Rule-based models describe clusters through a set of flat logical conditions rather than hierarchical decision paths.   In this example, the three clusters can be expressed as parallel rules operating at the same logical level:  samples satisfying $F_1 < 2.5$ and $F_2 < 2.5$ form Cluster~1;  those with $F_1 > 3$ and $F_2 > 2$ form Cluster~2;  and those with $F_1 < 2.5$ and $F_2 > 3$ form Cluster~3.  Each rule provides a self-contained, human-readable statement specifying the feature-value combinations that define cluster membership.  Compared with the decision tree, rule-based models establish direct logical relations between features and clusters.

\item \textbf{Prototype.} Prototype-based models define each cluster through a representative exemplar. In this example, three prototypes $\mathbf{p}_1$, $\mathbf{p}_2$, and $\mathbf{p}_3$ represent the three clusters, each with its own influence region in the feature space. For instance, samples near $\mathbf{p}_1$ at $(F_1=1, F_2=1)$ form Cluster~1, those around $\mathbf{p}_2$ at $(F_1=4.5, F_2=4.5)$ form Cluster~2, and those surrounding $\mathbf{p}_3$ at $(F_1=1, F_2=5)$ form Cluster~3. Interpretability arises from how each prototype locally captures its cluster while all prototypes collectively outline the global cluster structure. Unlike hierarchical or rule-based models relying on logical structure, prototype-based approaches use exemplars as compact, interpretable representations of clusters.

\item \textbf{Convex polyhedral.} Convex polyhedral models represent each cluster as a convex region enclosed by a set of bounding planes. In this example, three convex regions (rectangular in shape) are formed in the $(F_1, F_2)$ space, each defined by the intersection of two half-planes. Samples falling within $-0.5 < F_1 < 2.5$ and $-0.75 < F_2 < 2.75$ form Cluster~1; those within $2.75 < F_1 < 6$ and $2.5 < F_2 < 6$ form Cluster~2; and those within $-1 < F_1 < 3.25$ and $1.8 < F_2 < 5.5$ form Cluster~3. Interpretability arises from the explicit half-space boundaries that locally define each cluster's convex region, while all convex polyhedra collectively establish a coherent geometric representation of the global cluster structure.

\end{itemize}	

\textbf{\textit{Key differences.}} In terms of explanation, decision-tree models yield hierarchical paths composed of sequential decisions; rule-based models provide flat logical conditions (conjunctions or disjunctions) without hierarchy; prototype-based models explain clustering through similarity to representative exemplars; and convex polyhedral models provide explanations via explicit geometric constraints formed by half-space intersections. All four approaches produce human-readable descriptions but differ in both their representational form (path, flat rule, exemplar, or hyperplane) and the aspect they make explicit, including decision order, logical structure, representativeness, and geometric boundary.

\textbf{(3) Interpretability level.}
Existing methods can be categorized into \textit{model-level} and \textit{feature-level} interpretability based on their degree of explainability. While most of the methods discussed in this paper focus on designing interpretable models to obtain clustering results or fitting the results of third-party algorithms, some methods also emphasize the extraction of interpretable features from complex data, or the investigation of the relationships between specific clusters and their associated features, thereby enhancing interpretability.

\textbf{(4) Data modality.} Finally, methods can also be classified by the nature of data they are designed to process. Common data modalities include tabular data (numerical, categorical, or mixed), sequential data (e.g., discrete sequences or time series), as well as formats such as images, text, and graphs. The appropriate interpretable model and explanatory form often depend on the structural characteristics of the underlying data.

Overall, these four criteria provide a structured taxonomy that systematizes existing interpretable clustering methods, informs future methodological development, and establishes a conceptual bridge to the taxonomies of supervised eXplainable AI (XAI).

\section{Conceptual correspondence between interpretable clustering and supervised XAI}
\label{sec:correspondence}

\begin{table}[t]
	\centering
	\caption{Conceptual correspondence between supervised explainable AI (XAI) and interpretable clustering.}
	\label{tab:xai-conceptual}
	\small
	\begin{tabular}{p{0.22\linewidth} p{0.36\linewidth} p{0.36\linewidth}}
		\toprule
		\textbf{Taxonomy in supervised XAI} & \textbf{Meaning in supervised learning} & \textbf{Correspondence in interpretable clustering (unsupervised setting)} \\ 
		\midrule
		\textbf{Learning objective}~\cite{Guidotti2018} &  
		Predict labeled outcomes with explainable reasoning faithful to both labels and features. &  
		Discover latent structures or groupings in unlabeled data while providing human-understandable rationales for the resulting clusters. \\ 
		\addlinespace
		\textbf{Intrinsic (ante-hoc) vs.\ post-hoc}~\cite{Burkart2021} &  
		Intrinsic models are interpretable by design (e.g., decision trees, linear models), while post-hoc methods generate explanations for already-trained black-box models. &  
		In-clustering methods embed interpretability directly into the clustering process, whereas post-clustering methods derive explanations for pre-computed cluster partitions. \\  
		\addlinespace
		\textbf{Model-specific vs.\ model-agnostic}~\cite{Adadi2018} &  
		Model-specific explanations depend on the internal mechanisms or parameters of a particular model, while model-agnostic approaches can be applied to any model by analyzing its input-output behavior. &  
		All in-clustering methods are model-specific, as their interpretability stems from the internal structure of the clustering model itself, whereas post-clustering approaches can employ model-agnostic surrogate explainers to interpret black-box clustering outcomes. \\
		\addlinespace
		\textbf{Global vs.\ local explanation}~\cite{Bodria2023} &  
		Global explanations describe the overall decision logic of the model; local explanations clarify individual predictions. &  
		Cluster-level explanations correspond to global understanding of group-level patterns, whereas instance-level explanations clarify why a specific sample belongs to a cluster. \\ 
		\addlinespace
		\textbf{Explanation form}~\cite{Longo2024} &  
		Typical forms include decision trees, rules, prototypes, feature attributions, and counterfactuals. &  
		Analogous interpretable forms include decision trees, rules, prototypes, geometric boundaries, and data-dependent descriptions. \\ 
		\addlinespace
		\textbf{Evaluation criteria}~\cite{Pawlicki2024} &  
		Faithfulness, fidelity, stability, simplicity, and completeness, focusing on how well explanations align with model behavior and human reasoning. &
		Coverage and uniqueness of cluster explanations, rule or path simplicity, prototype sparsity, geometrically intuitive cluster boundaries (e.g., axis-aligned or visually interpretable regions). \\ 
		\bottomrule
	\end{tabular}
\end{table}

Recent surveys in supervised XAI have proposed comprehensive taxonomies that categorize explanation techniques according to their position in the learning pipeline, their dependency on model internals, and the form or scope of the explanations. Representative works include Adadi and Berrada~\cite{Adadi2018}, Guidotti et al.~\cite{Guidotti2018}, Gunning et al.~\cite{Gunning2019}, Burkart and Huber~\cite{Burkart2021}, and Bodria et al.~\cite{Bodria2023}, along with more recent efforts introducing refined evaluation frameworks and new interpretability paradigms~\cite{Chander2025,Moreira2025,Hossain2025,Bilal2025}. These studies have largely focused on predictive models, where the objective is to explain the mapping from input features to labeled outcomes. However, in unsupervised settings, no ground-truth labels exist, and the interpretive goal shifts from explaining predictions to elucidating the formation and meaning of discovered clusters. To highlight this conceptual distinction, we establish a correspondence between supervised XAI and interpretable clustering, illustrating how interpretability principles can be reformulated when supervision is absent.

The correspondence summarized in Table~\ref{tab:xai-conceptual} demonstrates that the foundational taxonomies of supervised XAI, originally developed for labeled prediction tasks, can be extended to unsupervised learning through careful reinterpretation of objectives and evaluation principles. For example, the distinction between intrinsic and post-hoc explanations parallels the in- and post-clustering categorization discussed in Section~\ref{sec:taxonomy}. Similarly, explanation forms such as trees, rules, and prototypes have direct counterparts in interpretable clustering, as illustrated in Fig.~\ref{models}.

Despite these conceptual parallels, interpretable clustering faces unique challenges that are not directly addressed in supervised XAI. The absence of ground-truth labels makes it difficult to validate explanations against reference outputs, shifting the emphasis toward internal consistency, statistical significance, and domain plausibility. These characteristics underscore the need for specialized evaluation metrics and flexible explanatory frameworks, in which in-clustering and post-clustering methods are often intertwined. Overall, this conceptual bridge illustrates how established XAI principles can inform interpretable clustering while highlighting the new methodological opportunities emerging in unsupervised interpretability research.

\section{Interpretable pre-clustering methods}
\label{sec:pre}

While most research on interpretable clustering focuses on generating transparent clustering outcomes, the interpretability of input features also plays a crucial role in ensuring understandable and trustworthy results. Existing interpretable pre-clustering methods aim to enhance feature interpretability at the data preparation stage and can be broadly categorized into two complementary perspectives: (1) \textit{interpretable feature extraction}, which derives human-understandable representations from complex data, and (2) \textit{interpretable feature selection}, which identifies compact and semantically meaningful subsets of features. Although both topics have been widely studied in the field of machine learning, they have rarely been systematically investigated in connection with subsequent clustering tasks.

From the perspective of feature extraction~\cite{Boniol2025}, studies have sought to obtain informative and human-understandable representations from complex data before clustering. Bonifati et al.~\cite{bonifati2022time2feat} proposed \textit{Time2Feat} to extract intra- and inter-signal features from multivariate time series using interpretable metrics and dimension reduction via Principal Feature Analysis or user annotation. Salles et al.~\cite{salles2024interpret3c} employed adaptive gating with Gumbel-SoftMax sampling to identify instance-relevant features that guide clustering, while Kang et al.~\cite{kang2024hyperspectral} introduced an interpretable hyperspectral band-selection algorithm based on Gestalt principles, aligning extracted representations with human visual perception to enhance the transparency of clustering outcomes.

From the perspective of feature selection, methods focus on identifying compact and discriminative feature subsets that preserve clustering accuracy while improving interpretability. Svirsky et al.~\cite{Svirsky2024} trained self-supervised local gates to produce instance-specific sparse feature sets, revealing which attributes drive each cluster assignment. Effenberger et al.~\cite{effenberger2021interpretable} applied a greedy feature selection strategy based on occurrence frequency and Jaccard similarity, generating concise and meaningful feature subsets that clarify the logic behind cluster formation.

\section{Interpretable in-clustering methods}
\label{sec:in}

Interpretable in-clustering methods constitute the core of interpretable clustering research, as they embed interpretability directly within the clustering process rather than applying it before or after clustering. In such methods, interpretability is treated as an explicit optimization objective jointly considered with traditional clustering criteria such as the within-cluster sum of squared errors (SSE). Some approaches formulate it as a multi-objective optimization~\cite{Carrizosa2023}, while others impose it as a structural regularization term~\cite{Hwang2023}.  

\textbf{\textit{Clarifying methodological distinctions.}}  
In-clustering methods are sometimes confused with pre- or post-clustering approaches because of when interpretability is introduced. Two criteria help clarify their methodological scope:

\begin{itemize}
	\item \textit{Dependence on third-party algorithms.}  
	This distinguishes in-clustering methods from post-clustering ones.  
	In-clustering integrates interpretability within the clustering objective, either by inducing clusters through interpretable models (e.g., tree-based clustering) or by jointly optimizing interpretability with standard costs, without relying on externally generated cluster labels~\cite{Gabidolla2022}.  
	In contrast, post-clustering methods explain reference partitions obtained from black-box algorithms.  
	For example, both~\cite{Laber2023} and~\cite{Hwang2023} optimize tree-based interpretability, yet the former fits a tree to fixed clustering results (post-clustering), whereas the latter jointly optimizes clustering and interpretability (in-clustering).  
	Thus, in-clustering emphasizes exploratory clustering guided by interpretability, producing clusters directly through transparent models~\cite{Rudin2019}.
	\item \textit{Interpretability during the clustering process.} This criterion distinguishes in-clustering from pre-clustering methods. Pre-clustering enhances feature interpretability before clustering, whereas in-clustering jointly learns interpretable features and cluster structures. For tabular data~\cite{Plant2011,Jiang2025a}, interpretability is straightforward through thresholds or categorical inclusion.  For more complex data such as networks~\cite{Pool2014,Sadler2022}, images~\cite{Wan2026}, and sequences, features lack clear semantics; thus, in-clustering methods aim to discover clusters and meaningful representations simultaneously. In networks, concise node descriptors are extracted for interpretable community detection~\cite{Atzmueller2016}; in images, semantic tags are identified via descriptive clustering~\cite{Dao2018}; and deep models improve interpretability through latent representation learning~\cite{Zhang2021,Pan2024}. For sequential data, vectorization often obscures meaning, motivating discriminative sequential pattern mining~\cite{Dong2025}. Some models integrate feature interpretability directly into clustering objectives: Kim et al.~\cite{Kim2015} group binary dimensions into logic-based interpretable sets, while Huang et al.~\cite{Huang2021} jointly optimize feature selection and clustering via a deep $K$-parallel auto-reconstructive framework with graph Laplacian regularization.
\end{itemize}

After clarifying these distinctions, the following subsections review representative in-clustering approaches, focusing on how interpretability objectives are embedded within clustering algorithms and realized through various interpretable models.

\subsection{Decision tree-based methods}
The decision tree model is widely recognized as an interpretable model in machine learning~\cite{Lundberg2020} and is commonly used for classification and regression tasks. Its interpretability stems from the recursive, hierarchical splitting of data based on feature values to generate intermediate results, and the final output is traceable through the feature values used in the splits. Instances are distributed to different leaf nodes (clusters) determined by specific splitting points according to certain criteria, following a clear, transparent path from the root node (representing the whole dataset) down through the branch nodes, which is easily understood by end users.

Early attempts to apply decision trees to clustering can be found in~\cite{Liu2000}, where uniformly distributed synthetic data were introduced as auxiliary data to build a standard (supervised) decision tree. This approach aimed to maximize the separation between the original data and the synthetic data by modifying the standard splitting criterion, such as information gain. Although this method used binary splits, which are relatively easy to understand, the reliance on data generation introduced additional assumptions, making it difficult to claim that the splits were truly interpretable. In contrast, \cite{Basak2005} developed an unsupervised decision tree directly based on the original features. The authors proposed four different measures for selecting the most appropriate feature and two algorithms for splitting data at each branch node. However, to select a candidate splitting point for calculating these measures, preliminary steps were required to divide the numerical feature domain into intervals~\cite{Chen2006,GutierrezRodriguez2015}. A simpler splitting criterion and a more intuitive algorithmic framework is presented in~\cite{Fraiman2013} with the introduction of CUBT, which was further extended to categorical data in~\cite{Ghattas2017}. CUBT adopts a general approach similar to CART~\cite{Krzywinski2017}, involving three steps: maximal tree construction, followed by pruning and merging to simplify the tree structure. This unsupervised decision tree-based clustering model was also extended to the interpretable fuzzy clustering domain in~\cite{Jiao2022}, where fuzzy splitting at branch nodes was used to grow the initial tree, followed by merging similar clusters to create a more compact tree structure.

The aforementioned unsupervised decision tree-based models adopt a top-down approach~\cite{Blockeel1998,Guidotti2023}, where all possible candidate splitting points are considered at the current branch node level, and criteria such as heterogeneity are calculated so that the tree grows greedily (\underline{greedy search}) based on the optimal splits passed down from the parent node. However, this type of algorithm lacks global guidance, meaning that each split is optimized locally rather than achieving a globally optimized solution across the entire dataset. 

Some advanced interpretable in-clustering methods that use decision trees leverage modern optimization techniques. These modern optimization techniques include, but are not limited to, Mixed-Integer linear Optimization (\underline{MIO}) techniques~\cite{Bertsimas2017} used in~\cite{Bertsimas2021}, Tree Alternating Optimization (\underline{TAO}) techniques~\cite{CarreiraPerpinan2018} used in~\cite{Gabidolla2022}, and monotonic optimization techniques such as the Branch-Reduce-and-Bound (\underline{BRB}) algorithm~\cite{Hellings2012} used in~\cite{Hwang2023}. These methods are designed to construct globally optimal clustering trees by explicitly optimizing a well-defined objective function applied to the entire dataset. Unlike traditional top-down approaches, these  methods directly establish a relationship between the instances assigned to different leaf nodes (clusters) and the interpretability objective, which is explicitly encoded in the objective function. These methods express interpretability in a more quantitative and formalized manner, often by specifying tree structural metrics~\cite{Piltaver2016} (e.g., the number of leaf nodes), where a smaller number of leaf nodes (\underline{nLeaf}), as used in~\cite{Gabidolla2022, Hwang2023}, typically indicates lower tree complexity and, correspondingly, better interpretability. Building on this global optimization framework, some interpretable fuzzy clustering algorithms are  presented as well. For example, \cite{Good2023} employs kernel density decision trees (KDDTs) for constructing fuzzy decision trees using an alternating optimization strategy, while \cite{Cohen2023} incorporates a soft (probabilistic) version of the split in their objective function and obtains the optimal split via a Constrained Continuous Optimization Model.

\subsection{Rule-based methods}
The process of mining an optimal rule set to derive a specific cluster is often inspired by the field of pattern mining~\cite{Han2012}. To ensure that different rule sets effectively correspond to their respective clusters, the rule set typically exhibits two key characteristics~\cite{Guilbert2025}: (1) frequency (meaningful), indicating that the rule set should cover as many samples within its corresponding cluster (true positives) as possible, and (2) discriminative power (unique), meaning that the rule set should minimize the number of samples mistakenly covered by other clusters (false positives).

To obtain a rule set for the purpose of interpretable clustering, a common approach is to start by quantifying interpretability based on how well a rule covers a specific cluster. For example, as demonstrated in~\cite{Saisubramanian2020}, an interpretability score is defined to assess a feature value's relevance to a cluster by considering the fraction of samples within the cluster that share that feature value. Given all candidate rules or rule sets (e.g., generated using frequent pattern mining), these  methods aim to derive clusters that maximize the interpretability score while simultaneously optimizing cluster quality. Since interpretability objectives often conflict with cluster quality, existing methods typically incorporate the interpretability score as a user-specified bound to balance interpretability and cluster quality, alongside standard clustering objectives. The method in~\cite{Carrizosa2023} introduces two explainability criteria for each rule set associated with a cluster: one similar to~\cite{Saisubramanian2020}, and another that considers the distinctiveness of the rule set, meaning how few samples it covers outside the associated cluster. Optimizing these two explainability objectives, together with cluster quality measures, is formulated into a multi-objective Mixed-Integer linear Optimization problem (\underline{multi-MIO}). Furthermore, the method in~\cite{Carrizosa2023} considers the maximum rule set length (\underline{lenRule}), i.e., the number of feature values in the combination, as a constraint, ensuring that the created clusters are more interpretable by being represented through concise rules.

Other interpretable rule-based methods may be customized, where the meaning of the rules is no longer based solely on feature values. For instance, in document datasets~\cite{Balachandran2009}, the rules may take different forms. Methods such as those in the field of fuzzy rule-based clustering~\cite{Gu2024}, have been summarized in the survey~\cite{yang2021survey}.

\subsection{Other methods}
In addition to the two widely used interpretable models mentioned above, other interpretable in-clustering methods create clusters or determine cluster membership based on representative elements~\cite{Chen2024a,Sabbatini2023}, which can generally be categorized as boundary-based or centroid-like approaches. However, for these representative elements to be interpretable, certain properties need to be maintained. The following is a brief overview of these approaches.

\textit{Convex-polyhedral}: These methods constrain the cluster boundaries to be axis-parallel (rectangular) in the feature space, as in the method proposed in~\cite{Chen2016}, which designs a Probabilistic Discriminative Model (\underline{PDM}) to define such clusters. More generally, they may use hyperplanes that allow for diagonal boundaries~\cite{Lawless2022} to more accurately represent a cluster.

In either case, the goal is to create clusters with fewer feature values, incorporating these as interpretability constraints within the standard clustering objective function. For instance, \cite{Lawless2022} uses a Mixed-Integer nonlinear Optimization (\underline{nonlinear-MIO}) programming formulation to jointly identify clusters and define polytopes. For axis-parallel boundaries, a single feature value is used per dimension, while diagonal boundaries rely on linear combinations of feature values. Although diagonal boundaries have greater power to distinguish different clusters, they are less interpretable due to their increased complexity compared to simpler axis-parallel boundaries.

\textit{Prototype (exemplar)}: In datasets where the original features are non-interpretable and difficult to understand, such as with images~\cite{Chen2024} and text~\cite{DiazRodriguez2026}, especially when deep embeddings are used, recent work on interpretable in-clustering via exemplars has found that seeking high-level centroids can be useful for characterizing clusters and facilitating visualization. For example, \cite{Davidson2024} tackles the challenging problem of finding the minimum number of exemplars (\underline{nExemplar}) without prior specification. Additionally, \cite{Pan2024} proposes a new end-to-end framework designed to enhance scalability for larger datasets, making exemplar-based clustering more practical for real-world applications.

\begin{table*}[t]
	\centering
	\caption{Summary of various interpretable in-clustering methods, each listing the representative reference and corresponding criteria.}
	\resizebox{\linewidth}{!}{%
	\begin{tabular}{ccccc}
		\toprule
		\textbf{Interpretable} & \textbf{Representative} & \textbf{Optimization} & \textbf{Interpretability-related} & \textbf{Axis-parallel} \\
		\textbf{model} & \textbf{reference} & \textbf{approach} & \textbf{structural metrics}  & \textbf{partitioning} \\
		\midrule
		\multirow{4}{*}{Decision Tree} & \cite{Fraiman2013} & greedy search & / & Yes \\
		& \cite{Bertsimas2021} & MIO & / & Yes \\
		& \cite{Hwang2023} & BRB & nLeaf & Yes \\
		& \cite{Gabidolla2022} & TAO & nLeaf & No \\
		\midrule
		\multirow{2}{*}{Rules} & \cite{Saisubramanian2020} & greedy search & / & Yes \\
		& \cite{Carrizosa2023} & multi-MIO & lenRule & Yes \\
		\midrule
		\multirow{2}{*}{Convex-polyhedral} & \cite{Chen2016} & PDM & / & Yes \\
		& \cite{Lawless2022} & nonlinear-MIO & / & No \\
		\midrule
		\multirow{2}{*}{Prototype} & \cite{Pan2024} & stochastic gradient & / & No \\
		& \cite{Davidson2024} & greedy search & nExemplar & No \\
		\bottomrule
	\end{tabular}}
	\label{tab:in-clustering}
\end{table*}

\subsection{Summary}
Various interpretable models have been developed for in-clustering methods, as summarized in Table~\ref{tab:in-clustering}. Currently, most existing approaches in the literature either build upon these representative methods or follow similar principles that can be subsumed under them. These models consistently treat interpretability as a first-class objective, on par with clustering quality, incorporating it as an optimization target either directly or indirectly, depending on the model type. For instance, tree-based models often prioritize reducing the number of branch or leaf nodes, rule-based models focus on shorter rules, and geometric representation models, such as prototype-based models, aim to minimize the number of exemplars. More refined structural parameters as optimization targets require further research. For example, in literature~\cite{Laber2023}, tree depth is considered an optimization target; however, this approach, designed to explain a given reference clustering result, belongs to post-clustering methods.

There is often a trade-off between interpretability and clustering quality, where enhancing one may diminish the other. This frequently addressed challenge could be less daunting in post-clustering methods, which only need to focus on one direction, specifically fitting given clustering results. In contrast, in-clustering methods must account for the simultaneous pursuit of both objectives. A critical research direction for in-clustering methods is to balance these objectives while ensuring scalability for real-world data. As shown in Fig.~\ref{models}, several interpretable models cannot perfectly predict all samples with respect to their clusters. While standard decision tree models generate partitions aligned with coordinate axes, more flexible oblique decision trees~\cite{Gabidolla2022} can improve clustering performance. Similarly, convex-polyhedral approaches can benefit from allowing diagonal boundaries~\cite{Lawless2022}, not limited to axis-parallel rectangles, provided they remain convex. Further research is needed to design new interpretable models that can effectively handle complex data.

\section{Interpretable post-clustering methods}
\label{sec:post}
Post-modeling interpretability is a crucial aspect of interpretable learning, focusing on elucidating the reasoning behind decisions made by black-box models. In the context of clustering, interpretable post-clustering refers to the use of interpretable models, such as decision trees, to closely approximate existing clustering results (also known as reference clustering results). This means that the labels assigned to samples by the interpretable model should align as closely as possible with the original results. This kind of method aids in understanding why certain samples are assigned to specific clusters, thereby fostering trust in black-box models. In the following subsections, we will categorize existing interpretable post-clustering methods based on different interpretable models.

\subsection{Decision tree-based methods}

Decision trees are the most widely used interpretable models for post-clustering analysis. In a decision tree, each internal node splits the samples it contains into different groups based on predefined criteria. The $k$ leaf nodes (not necessarily the ground-truth cluster number) correspond to the $k$ clusters in the reference clustering results. Each cluster assignment can be interpreted by the path leading to its respective leaf node.

In decsion tree-based post-clustering methods, the closer the clustering results obtained by the constructed decision tree are to the reference clustering results, the better its interpretability performance~\cite{Argov2025}. This metric is often defined in existing research as ``the price of interpretability" \cite{moshkovitz2020explainable},  which is the ratio of the cost of the explainable clustering to the cost of an optimal clustering (e.g., $k$-means/medians). Therefore, the goal is typically to build a decision tree $T$ such that 
$cost(T$) is not too large compared to the optimal $k$-means/medians cost. Specifically, an algorithm is said to have an $x$-approximation guarantee if the cost of the tree is at most $x$ times the optimal cost, i.e., if the algorithm returns a threshold tree $T$, then we have $cost(T) < x\cdot cost(opt)$.

Research on the quality of decison tree constructed by interpretable post-clustering methods  began with the work of Moshkovitz et al. \cite{moshkovitz2020explainable}. They develop decision trees using a greedy approach that aims to minimize the number of errors at each split (i.e., the number of points separated from their corresponding reference cluster centers), stopping when the tree reaches $k$ leaf nodes. This method achieves an $O(k)$ approximation for the optimal $k$-medians and an $O(k^2)$ approximation to the optimal $k$-means. 
Laber et al. \cite{laber2021price} improve the approximation, achieving an  $O(d \log k)$ approximation for optimal $k$-medians and an $O(k d \log k)$ approximation for the optimal $k$-means. They accomplish this by firstly constructing $d$ decision trees, where $d$ is the number of dimensions in the data, then utilize these trees  to build the final decision tree.  The feature for splitting a node within the final decision tree is chosen based on the dimension with the maximum range among the centers contained in the current node. The specific feature value is associated with the node in the corresponding dimension's decision tree, which is the least common ancestor (LCA) of the set of reference centers that reach the current node.
Makarychev et al. \cite{makarychev2021near} take a different approach by choosing splitting features and values that differentiate centers with greater distances within each node in a relatively random manner.  This results in an $O(\log k \log \log k)$ approximation for the optimal $k$-medians and an $O(k \log k \log \log k)$ approximation for the optimal $k$-means, with $\log \log k$ denoting the iterated logarithm $\log(\log k)$.  
In the decision tree constructed in \cite{gamlath2021nearly}, the choice of cuts at each split node is entirely random, as long as it can separate different reference centers into different child nodes. It has been proven that this method can achieve an $O(\log^2 k)$ approximation for the optimal $k$-medians and an $O(k \log^2 k)$ approximation for the optimal  $k$-means. 
Recentlty, Esfandiari et al. \cite{esfandiari2022almost}  focus on determining the maximum and minimum values of the reference centers along each dimension, sorting these values, and then sampling a split point that effectively separates the reference centers. Their method achieves an  $O(\log k \log \log k)$ approximation for the optimal $k$-medians and an $O(k \log k)$ approximation for $k$-means. 
Several methods have been proposed to independently provide near-optimal algorithms for $k$-means or $k$-medians \cite{charikar2022near,byrka2017improved,makarychev2024random}, which will not be elaborated upon here.

Unlike focusing on improving a decision tree model's ability to provide an approximation guarantee for optimal clustering results, Frost et al. \cite{frost2020exkmc} adopt the method from \cite{Laber2023} to build a tree with $k$ leaf nodes and then use a new surrogate cost to greedily expand the tree to $k'>k$ leaves,  proving that as $k'$  increases, the surrogate cost is non-increasing. This approach reduces clustering cost while providing a flexible trade-off between interpretability and accuracy. Laber et al. \cite{Laber2023} focus on building decision trees that yield short explanations (i.e., trees with smaller depth) for the clusters of the partition while still inducing good partitions in terms of the $k$-means cost function. Additionally, they propose two structural metrics for measuring interpretability: Weighted Average Depth (\underline{WAD}), which weighs the depth of each leaf by the number of samples in its associated cluster, and Weighted Average Explanation Size (WAES), a variation of WAD. Inspired by robustness studies, Bandyapadhyay et al. \cite{bandyapadhyay2023find} explore constructing a decision tree by removing the fewest points necessary to match the reference clustering results exactly, where interpretability is measured by the number of points removed.

\subsection{Rule-based methods}
Distinct from decision trees, interpretable post-clustering models constructed using if-then rules do not involve hierarchical relationships. Their explanations for clusters are relatively concise and intuitive, providing a set of rules to describe the samples within a cluster. To our knowledge, despite the fact that if-then rules have become widely accepted as interpretable models and have been studied considerably, most rule-based interpretable clustering methods focus on extracting rules from data to form clusters. Consequently, there is limited research on post-clustering methods that generate rules and provide explanations for clusters that have already been formed.

Carrizosa et al.~\cite{Carrizosa2023} explain clusters with the objective of maximizing the total number of true positive cases (i.e., the number of samples within the cluster that satisfy the explanation) and minimizing the total number of false positive cases (i.e., the number of individuals outside the cluster that satisfy the explanation). Additionally, the length of the rules is constrained to ensure strong interpretability.

De Weerdt et al. \cite{de2014secpi} investigate the search for explanations for event logs by first generating feature sets from the data and then applying a best-first search procedure with pruning to construct the set of explanations. Through an iterative process, they continuously enhance the accuracy and conciseness of the explanations for the instances. Building on this work, Koninck et al. \cite{de2017explaining} mine concise rules for each individual instance from a black-box support vector machine (SVM) model and discuss and evaluate different alternative feature sets that can be used as inputs for explanatory techniques.

Ofek et al.~\cite{Ofek2025} introduced Cluster-Explorer, a post-hoc framework that explains black-box clustering pipelines through compact rule sets. The method formulates explanation generation as a multi-objective optimization over coverage, separation error, and conciseness, which is efficiently solved using a generalized frequent itemset mining (gFIM)-based search. By representing predicates as items and applying Pareto pruning, the algorithm derives minimal-length rule sets that maximize cluster coverage while minimizing overlap across clusters.

\subsection{Other methods}
Besides the aforementioned decision trees and if-then rules, several other interpretable models have been used in literature to explain existing clustering results. Given their limited number, we will not review each interpretable model individually but rather provide an overall summary here.

\textit{Prototype}. Carrizosa et al. \cite{carrizosa2022interpreting} proposed a method for using prototypes to explain each cluster. A prototype is an individual that serves as a representative example of its cluster, defined by its minimal dissimilarity to other individuals within the same cluster.  In their approach, they solve a bi-objective optimization problem to identify these prototypes. This problem aims to maximize the number of true positive cases within each cluster while minimizing the number of false positive cases in other clusters.

\textit{Convex polyhedral}. In \cite{lawless2023cluster}, a polyhedron is constructed around each cluster to serve as its explanation. Each polyhedron is formed by intersecting a limited number of half-spaces (\underline{nHalfspace}). The authors formulate the polyhedral description problem as an integer program, where variables correspond to candidate half-spaces for the polyhedral description of the clusters. Additionally, they present a \underline{column generation} approach to efficiently search through the candidate half-spaces. Chen et al. \cite{chen2023explanation} propose using a hypercube coverage model to explain clustering results. This model incorporates two objective functions: the number of hypercubes (\underline{nHypercube}) and the compactness of instances. A \underline{heuristic search} method (NSGA-II) is employed to identify a set of non-dominated solutions, defining an ideal point to determine the most suitable solution, whereby each cluster is covered by as few hypercubes as possible.

\textit{Description}. Davidson et al. \cite{davidson2018cluster} introduce the cluster description problem, where each data point is associated with a set of descriptions from a discrete set. The objective is to find a set of non-overlapping descriptions for each cluster that covers every instance within the cluster. The proposed method allows for the specification of the maximum number of descriptions per cluster and the maximum number of clusters that any two descriptions can jointly cover.

\begin{table*}[t]
	\centering
	\caption{Summary of various interpretable post-clustering methods, each listing the representative reference and corresponding criteria.}
	\resizebox{\linewidth}{!}{%
	\begin{tabular}{ccccc}
		\toprule
		\textbf{Interpretable} & \textbf{Representative} & \textbf{Optimization} & \textbf{Interpretability-related} & \textbf{Axis-parallel} \\
		\textbf{model} & \textbf{reference} & \textbf{approach} & \textbf{structural metrics}  & \textbf{partitioning} \\
		\midrule
		\multirow{2}{*}{Decision Tree} & \cite{moshkovitz2020explainable} & greedy search & / & Yes \\
		& \cite{Laber2023} & greedy search & WAD & Yes \\
		\midrule
		\multirow{2}{*}{Rules} & \cite{Carrizosa2023} & MIO & lenRule & Yes \\
		& \cite{Ofek2025} & gFIM-based & lenRule & Yes  \\
		\midrule
		\multirow{2}{*}{Convex-polyhedral} & \cite{lawless2023cluster} & column generation & nHalfspace & No \\
		& \cite{chen2023explanation} & heuristic search & nHypercube & Yes \\
		\midrule
		\multirow{1}{*}{Prototype} & \cite{carrizosa2022interpreting} & MIO & / & No \\
		\bottomrule
	\end{tabular}}
	\label{tab:post-clustering}
\end{table*}

\subsection{Summary}
Several representative interpretable post-clustering  methods are summarized in Table~\ref{tab:post-clustering}. Additionally, the following observations can be noted: firstly, most post-clustering research utilizes decision trees as interpretable models to explain clustering results. However, explanations derived from decision trees have certain drawbacks~\cite{Zhang2026}, such as the dependency of deep-layer decisions on shallow-layer decisions. Additionally, it is possible to consider using a hyperplane in a chosen number of dimensions instead of splitting along only one feature. Moreover, the choice of a suitable interpretable model may vary depending on the type of data; for instance, descriptions may be more appropriate for community analysis. Therefore, the post-clustering methods involving other interpretable models require further investigation.

Secondly, existing methods primarily focus on approximating the optimal clustering cost of reference clustering results using decision tree-based approaches, or aiming for interpretable models with high true positive rates and low false positive rates \cite{carrizosa2022interpreting,Carrizosa2023}. However, few methods emphasize the simplicity of explanations (except for \cite{Laber2023,Carrizosa2023}), which includes but is not limited to the depth of decision trees, the number of leaf nodes, and the length and quantity of rules. Thus, the balance between the accuracy and simplicity of interpretable models, as well as the quantification of interpretability metrics, remains an area for further research.

\section{Future directions}

To provide valuable insights for the future direction of this field, we have classified various interpretable clustering methods based on different aspects and further summarized key technical criteria for readers' reference, such as: (1) Optimization approaches, which illustrate how authors from various domains have formalized the interpretability challenges in clustering and the methods they have employed to solve these optimization problems, and (2) Interpretability-related structural metrics, which are crucial as they could potentially be utilized to evaluate the interpretability quality of novel methods, similar to how accuracy is used to assess clustering quality. The literature still lacks attention to a greater diversity of these structural metrics. We believe that researchers studying these different interpretable clustering methods can complement and enhance each other’s work. Moreover, methods from different clustering stages could be combined, as relying solely on a single-stage interpretable clustering method may be insufficient for complex and challenging application scenarios. This is particularly true in cases where obvious interpretable features do not exist, making it difficult to construct interpretable clustering algorithms. Additionally, research on interpretable clustering methods for intricate data, such as categorical data~\cite{Hu2025,Hu2025a,Chen2026}, discrete sequences~\cite{Dong2025,He2025,Liu2026}, time series~\cite{Huang2025,Schlegel2026}, network (graph)~\cite{Sun2025,Li2025,Sun2026}, and multi-view and multi-modal data~\cite{Jiang2025,Wang2025}, remains limited.

\section{Conclusion}
\label{conclusion}

This survey provides a comprehensive and systematic perspective on various interpretable clustering methods, highlighting both foundational research and the latest advancements in the field. It is the first to address the topic across the full lifecycle of clustering analysis, encompassing Pre-clustering, In-clustering, and Post-clustering stages. At each stage, relevant literature on interpretable clustering methods is reviewed. Primarily, this work aims to clearly define what interpretability means in the context of clustering and how it is embedded in commonly used interpretable models, such as decision trees, rules, prototypes, and convex polyhedral models. These models create interpretable clusters with elements that are understandable to human users and potentially enable these clustering results to be applied in high-risk domains, meeting essential prerequisites of transparency and trustworthiness~\cite{Miedema2026}. The effort to endow various clustering paradigms with explainability is still in its infancy~\cite{Ren2023,Lv2025,Miller2025,Souza2025,Li2026}.

\begin{acks}
	This work has been supported by  the National Natural Science Foundation of China under Grant No. 62472064.
\end{acks}

\bibliographystyle{ACM-Reference-Format}
\bibliography{refj}

\end{document}